\documentclass[10pt,twocolumn,letterpaper]{article}

\usepackage{graphicx}
\usepackage{color}

\usepackage{pifont}
\usepackage{multirow}
\usepackage{makecell}
\usepackage{stfloats}
\usepackage[linesnumbered,ruled,vlined]{algorithm2e}

\usepackage[pagenumbers]{iccv} 

\definecolor{iccvblue}{rgb}{0.21,0.49,0.74}
\usepackage[pagebackref,breaklinks,colorlinks,allcolors=iccvblue]{hyperref}

\def\paperID{3423} 
\def\confName{ICCV}
\def\confYear{2025}

\title{From Flat to Round: Redefining Brain Decoding with Surface-Based fMRI and Cortex Structure}

\author{\textbf{Sijin Yu}$^1$ ~~~ \textbf{Zijiao Chen}$^2$ ~~~ \textbf{Wenxuan Wu}$^1$ ~~~ \textbf{Shengxian Chen}$^1$ ~~~ \textbf{Zhongliang Liu}$^1$\\
\textbf{Jingxin Nie}$^3$ ~~~ \textbf{Xiaofen Xing}$^1$ ~~~ \textbf{Xiangmin Xu}$^{1,4}$ ~~~ \textbf{Xin Zhang}$^{\star~1}$\\
$^1$South China University of Technology\\
$^2$Stanford University\\
$^3$South China Normal University\\
$^4$Pazhou Lab\\
}

\begin{document}

\twocolumn[{
\renewcommand\twocolumn[1][]{#1}%
\maketitle

}]



\begin{abstract}

Reconstructing visual stimuli from human brain activity (\textit{e.g.}, fMRI) bridges neuroscience and computer vision by decoding neural representations.
However, existing methods often overlook critical brain structure-function relationships, flattening spatial information and neglecting individual anatomical variations.
To address these issues, we propose (1) a novel sphere tokenizer that explicitly models fMRI signals as spatially coherent 2D spherical data on the cortical surface; (2) integration of structural MRI (sMRI) data, enabling personalized encoding of individual anatomical variations; and (3) a positive-sample mixup strategy for efficiently leveraging multiple fMRI scans associated with the same visual stimulus. 
Collectively, these innovations enhance reconstruction accuracy, biological interpretability, and generalizability across individuals. 
Experiments demonstrate superior reconstruction performance compared to SOTA methods, highlighting the effectiveness and interpretability of our biologically informed approach.

\end{abstract}

\section{Introduction}
\begin{figure*}[t]
    \centering
    \includegraphics[width=\textwidth]{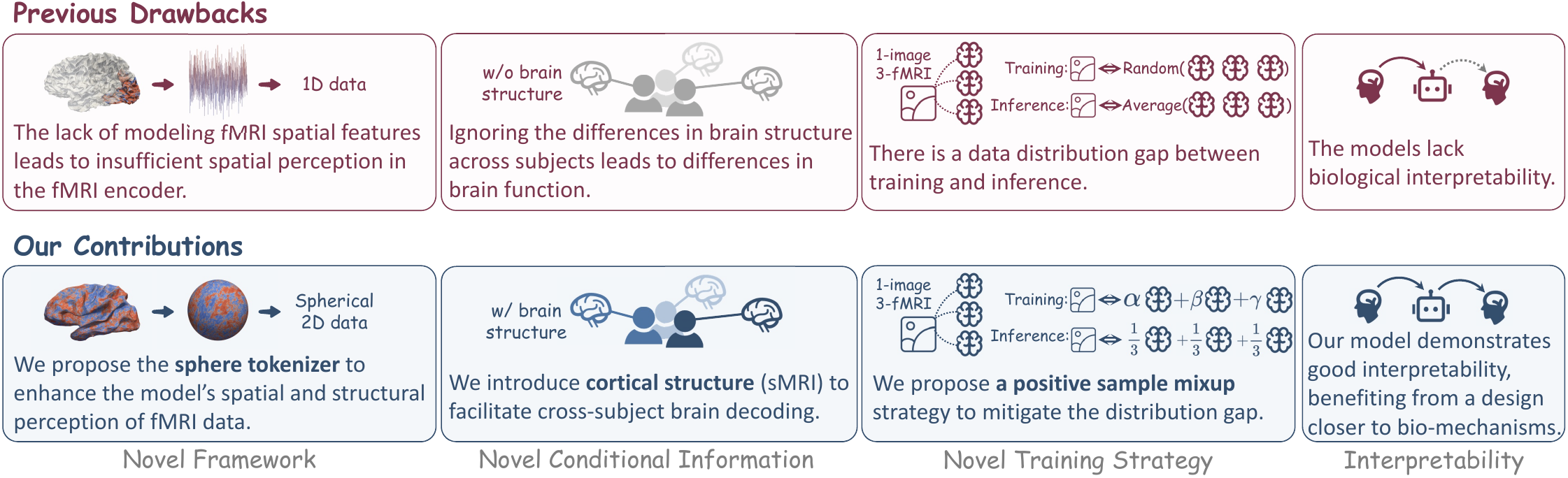}
    \caption{
    The drawbacks of previous work and the \textbf{contributions} of this paper.
    }
    \label{fig:contributions}
\end{figure*}

The human brain, shaped by evolution, can be viewed as a highly optimized, naturally ``pre-trained'' neural network. 
It encodes sensory stimuli such as visual and auditory signals from various organs into intricate patterns of neural activity. 
Brain decoding, the reverse of this process, aims to reconstruct sensory stimuli from recorded brain activity, typically measured using functional magnetic resonance imaging (fMRI).
This decoding holds significant implications for both neuroscience, by revealing how the brain represents sensory information, and brain-computer interfaces (BCIs), by translating neural signals into actionable outputs.

A pivotal area of brain decoding is fMRI-image reconstruction, which focuses on the exploration of the brain’s visual functions.
Over time, techniques have evolved from semantic category reconstruction \cite{cox2003functional, kamitani2005decoding, du2023decoding} to pixel-level reconstruction \cite{beliy2019voxels, shen2019deep}.
Currently, the most popular approach is to train an fMRI encoder \cite{takagi2023high, chen2023seeing, ozcelik2023natural, scotti2024reconstructing, sun2024contrast, xia2024dream, chen2023cinematic}, whose output fMRI embeddings are used as conditions to guide a diffusion model \cite{ho2020denoising, dhariwal2021diffusion, rombach2022high} for image reconstruction.
Following these approaches, fMRI-image reconstruction has expanded from single-subject to cross-subject applications \cite{wang2024mindbridge, quan2024psychometry, xia2025umbrae, huo2025neuropictor, shenneuro}.
Methods that encode fMRI signals and use them as conditions to guide diffusion models have achieved success.

However, existing methods share fundamental limitations stemming from the oversimplification of fMRI data as one-dimensional signals, disregarding critical spatial and structural properties. 
Specifically:
(1) fMRI signals arise from voxels located on the cerebral cortex, naturally forming spatial patterns on a two-dimensional non-Euclidean cortical surface \cite{fischl1999cortical}. 
Ignoring this spatial organization diminishes the richness of spatial information available.
(2) Structural brain differences among individuals significantly influence functional responses \cite{pang2023geometric, fotiadis2024structure}, resulting in varied neural responses to identical stimuli \cite{khosla2021cortical}. 
Current methods fail to effectively capture these structural variations, thus impairing performance in cross-subject generalization.

Based on this observation, we propose the Sphere Tokenizer, which processes fMRI signals into fMRI tokens that incorporate spatial structural information. 
These tokens are then fed into the fMRI encoder for encoding.
fMRI signals located on the cerebral cortex are mapped onto a standard sphere using FreeSurfer-based \cite{fischl2012freesurfer} method.
Spherical convolution \cite{jiang2019spherical, zhao2019spherical} is introduced to extract fMRI features. 
Similarly to planar convolution \cite{krizhevsky2012imagenet}, each voxel (pixel) only gathers information from its neighbors and updates its value.
In addition, unlike previous spherical convolution frameworks, we introduce structural information of the cerebral cortex and sphere positional embedding into the Sphere Tokenizer.
The cortical structure allows the model to perceive differences in brain structures across subjects, while the sphere positional embedding enables the model to account for functional pattern differences across different regions of the cortex.
To the best of our knowledge, we are the first to incorporate brain structure into the fMRI-image reconstruction task.

In addition, current fMRI-image datasets typically provide asymmetric matching samples, \textit{i.e.}, one image corresponds to multiple fMRI scans.
Previous work apply each individual fMRI scan for training, but employ the averaged fMRI scan for inference.
This creates a data distribution gap between training and inference.
To address this, we proposed a positive sample mixup strategy, where during training, we randomly mixup the corresponding fMRI scans to match the averaging process during inference.
\textbf{Contributions:}
we summarize the drawbacks of previous works and the contributions of this paper in Fig. \ref{fig:contributions}.

\section{Related Work}
\paragraph{fMRI-Image Reconstruction.}
Early works of vision brain decoding are represented by approaches using sparse linear regression \cite{horikawa2017generic}, VAE-based methods \cite{du2023decoding}, and GAN-based methods \cite{seeliger2018generative, ozcelik2022reconstruction, gu2024decoding}.
With the advancement of image generation models, brain decoding tasks have evolved from semantic classification to pixel level reconstruction.
With the development of multimodal technologies and diffusion models \cite{ho2020denoising, dhariwal2021diffusion, rombach2022high}, the current popular pipeline involves aligning fMRI with CLIP \cite{radford2021learning} and then using it as a condition for diffusion models to perform image reconstruction \cite{chen2023seeing, takagi2023high, scotti2024reconstructing, sun2024contrast, xia2024dream}.
Building on this, researchers have extended this pipeline to cross-subject scenarios \cite{wang2024mindbridge, quan2024psychometry, xia2025umbrae, huo2025neuropictor, shenneuro}.
These methods typically achieve cross-subject training by introducing subject-specific parameters into the model \cite{wang2024mindbridge, quan2024psychometry} or incorporating subject-independent tokens \cite{xia2025umbrae}.

\paragraph{fMRI Tokenizer.}
The tokenizer processes the raw fMRI data into an appropriate format and passes it to the model.
A simple approach is to treat fMRI as 1D data and then process it using a linear module \cite{takagi2023high} or as one-dimensional convolution \cite{chen2023seeing}.
The linear tokenizer and 1D convolution tokenizer simply flatten the fMRI data, without taking its spatial and structural information into account.
Therefore, the ROI embedder is proposed \cite{qian2023joint, quan2024psychometry}, which processes fMRI data from different brain regions separately, enabling the fMRI tokens to have better spatial representation capabilities.
Unlike the ROI-wise spatial representation of fMRI, our proposed sphere tokenizer provides voxel-wise spatial representation capability.

\section{Method}

\begin{figure*}[t]
    \centering
    \includegraphics[width=\textwidth]{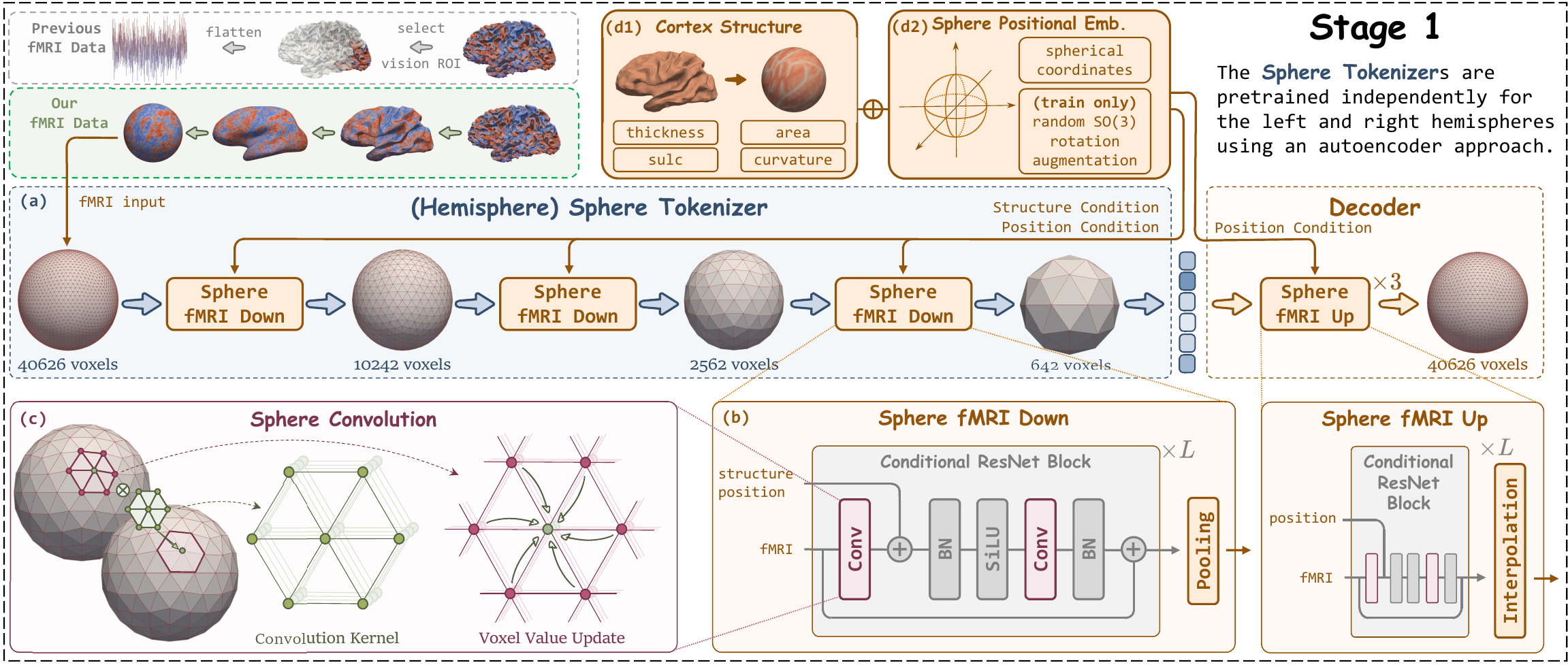}
    \caption{
    Illustration of sphere tokenizer (\textsection \ref{sec:method-sphere_tokenizer}).
    \textcolor[RGB]{52,78,109}{\textbf{(a)} [blue]} The tokenizer downsamples the fMRI data three times and maps it to fMRI tokens.
    \textcolor[RGB]{147,84,10}{\textbf{(b)} [brown, bottom]} Each downsampling layer consists of $L$ conditional ResNet blocks.
    \textcolor[RGB]{124,53,77}{\textbf{(c)} [purple]} The ResNet blocks are based on the sphere convolution, a 2D convolution on the sphere with a receptive field of one-hop neighbors.
    \textcolor[RGB]{147,84,10}{\textbf{(d)} [brown, top]} The conditional ResNet block has two types of conditional guidance: cortex structure \textcolor[RGB]{147,84,10}{(d1)} and sphere positional embeddings \textcolor[RGB]{147,84,10}{(d2)}.
    }
    \label{fig:sphere_tokenizer}
\end{figure*}

\subsection{Data Elaboration and Task Setup}

\paragraph{Dataset}

Our experiments build upon the Natural Scenes Dataset (NSD)~\cite{allen2022massive}, which contains large-scale fMRI--image pairs for eight healthy adult participants. The images are sampled from MS-COCO~\cite{lin2014microsoft}. Following standard practice \cite{scotti2024reconstructing, quan2024psychometry, wang2024mindbridge, shen2025neuro}, we focus on four participants (subj01, 02, 05, 07), each having viewed 10,000 images (scanned three times per image). Of these, 1,000 images are shared across all four participants, forming our \texttt{test} set. The remaining images are split into 8,500 for \texttt{train} and 500 for \texttt{val}. 
We use the fMRI data mapped to the fsaverage surface provided by NSD, then resample it onto the FreeSurfer \cite{fischl2012freesurfer} fsaverage6 standard mesh (40,962 $\times$ 2 voxels per subject). Further preprocessing details are in Appendix~\S\ref{appendix:data_preprocess}.

\paragraph{Task Setup and Notations.}

Our goal is to reconstruct an image $I$ from fMRI responses. Each image $I$ is paired with a COCO annotation $T$ and three corresponding fMRI scans $\{F_i^\texttt{L}, F_i^\texttt{R}\}$ ($i=1,2,3$), which represent left ($\texttt{L}$) and right ($\texttt{R}$) hemisphere activities. The model inputs $F_i^{\texttt{L}}, F_i^{\texttt{R}}\in\mathbb{R}^{n\times1}$ (where $n$ is the number of voxels on a hemisphere), and outputs a reconstructed image $\hat{I}$. 

To capture both spatial and structural variations in the cortex, we also introduce two additional conditions: cortical structure information $\{S^{\texttt{L}}, S^{\texttt{R}}\}$ and sphere positional embeddings $P$ (see \S\ref{sec:method-sphere_tokenizer}). These elements provide the tokenizer with explicit cues about individual structural differences and topological positions on the cortical surface.

\begin{figure*}[t]
    \centering
    \includegraphics[width=\textwidth]{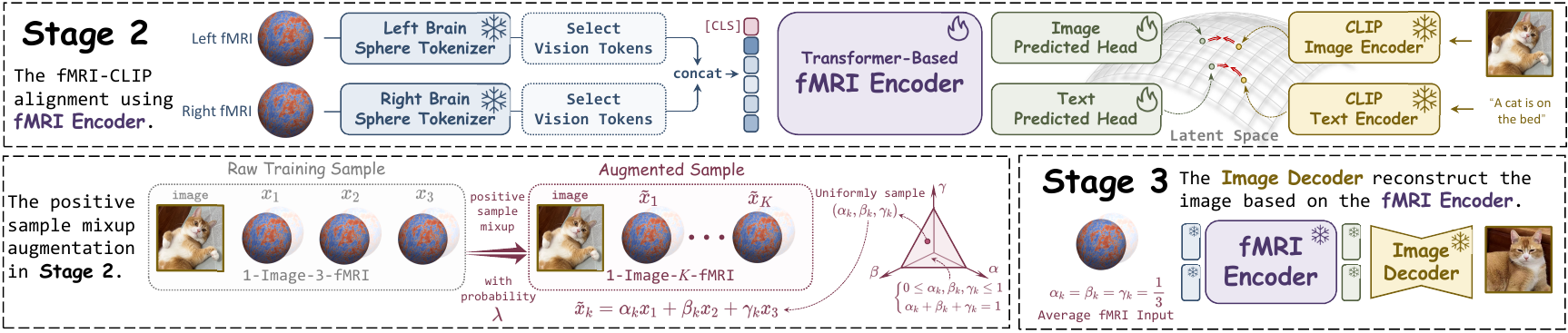}
    \caption{
    Illustration of the fMRI encoder (\S\ref{sec:method-fmri_image_text_alignment}) and the image reconstruction pipeline (\S\ref{sec:method-image_reconstruction}).
    \textbf{(top)} We select vision-relevant tokens from the sphere tokenizer output and feed them into the fMRI encoder, which predicts both image and text embeddings aligned with CLIP.
    \textbf{(bottom left)} Our positive sample mixup strategy: when selected, multiple fMRI scans of the same image are mixed with random weights to better approximate the averaging process used at inference.
    \textbf{(bottom right)} At test time, the three fMRI scans are averaged; the image decoder then reconstructs the final image from the aligned fMRI embedding. 
    }
    \label{fig:clip_and_inference}
\end{figure*}

\paragraph{Method Overview.}
\textbf{Stage 1}: A sphere tokenizer (\textsection \ref{sec:method-sphere_tokenizer}, Fig. \ref{fig:sphere_tokenizer}) first maps the raw fMRI data to the appropriate fMRI tokens.
The sphere tokenizer has strong spatial and structural perception due to (i) data interaction on a 2D spherical surface instead of the previous 1D space, (ii) the incorporation of individual cortical structure information $S^{\texttt{L}}, S^{\texttt{R}}$, and (iii) spherical position embedding $P$.
\textbf{Stage 2}: An fMRI encoder (\textsection \ref{sec:method-fmri_image_text_alignment}, Fig. \ref{fig:clip_and_inference}) takes the fMRI tokens output by the sphere tokenizer as input and aligns it with CLIP \cite{radford2021learning}.
\textbf{Stage 3}: An image decoder (\textsection \ref{sec:method-image_reconstruction}, Fig. \ref{fig:clip_and_inference}) reconstructs the image from the aligned fMRI embedding.

\subsection{Sphere Tokenizer}
\label{sec:method-sphere_tokenizer}

To leverage the inherent 2D structure of cortical data, we design the sphere tokenizer to project fMRI signals into fMRI tokens.
It encodes raw fMRI signals on a spherical surface, then reconstructs them at high resolution to ensure spatial fidelity. 
We train separate tokenizers for the left and right hemispheres (denoted $\mathcal{T}^\texttt{L}$, $\mathcal{T}^\texttt{R}$), given their roughly mirrored structural layout.

\paragraph{Overview of the Tokenizer.}
Focusing on the left hemisphere for illustration, the tokenizer $\mathcal{T}^\texttt{L}$ takes high-resolution fMRI data $F_i^\texttt{L}$ on fsaverage6 (40,962 voxels), progressively downsamples it three times until fsaverage3 (642 voxels), generating the \emph{fMRI token} $x^\texttt{L}$. A lightweight decoder then upsamples $x^\texttt{L}$ back to the fsaverage6 space to compute reconstruction losses (MSE and L1). Formally,
\begin{align}
    x^\texttt{L} = \mathcal{T}^{\texttt{L}}\bigl(F^\texttt{L} \mid S^\texttt{L}, P\bigr),
\end{align}
where $S^\texttt{L}$ is cortical structure information and $P$ is the sphere positional embedding.

\paragraph{Incorporating Brain Structure and Position.}
Prior research~\cite{pang2023geometric, fotiadis2024structure} shows that structural brain properties (\textit{e.g.}, thickness, curvature) strongly influence neural functions. Because no two brains are identical, we introduce structural MRI data to help the tokenizer separate function from individual anatomical variation. In parallel, we add a sphere positional embedding ($P$) based on 3D coordinates on the cortical surface. This helps the tokenizer distinguish signals arising from different cortical locations. 

\paragraph{Sphere Conditional ResNet Blocks.}
Each downsampling or upsampling stage stacks multiple conditional ResNet blocks, each containing two sphere convolutions~\cite{zhao2019spherical} that update voxel values using information from their one-hop neighbors on the spherical mesh. Between the two convolutions, we inject structural and positional features to condition the ResNet blocks. Specifically:
\begin{itemize}[leftmargin=1em]
    \item \textbf{Structure Condition} $S^\texttt{L} \in \mathbb R^{n\times 4}$: 
    We concatenate four cortical properties (thickness, surface area, sulcal depth, curvature) and resample them to match the current fMRI resolution in the ResNet blocks.
    \item \textbf{Positional Condition} $P \in \mathbb R^{n\times 3}$:
    We project each voxel’s 3D coordinates by an MLP, optionally applying random rotations to improve robustness to orientation. 
\end{itemize}
Together, these blocks yield \emph{fMRI tokens} that capture local spatial context (via sphere convolution) and individual cortical distinctions (via structural and positional conditions).

\subsection{fMRI-Image-Text Alignment}
\label{sec:method-fmri_image_text_alignment}

We apply a transformer-based \cite{vaswani2017attention} fMRI encoder ($\mathcal E_{\texttt{fMRI}}$), which takes fMRI tokens as input and aligns it with the CLIP \cite{radford2021learning} (image encoder: $\mathcal E_{\texttt{image}}$, text encoder: $\mathcal E_{\texttt{text}}$) embedding.
The input to $\mathcal E_{\texttt{fMRI}}$ is the fMRI vision tokens, which are based on the output of the pretrained sphere tokenizer.
The output of $\mathcal E_{\texttt{fMRI}}$ is the predicted image embedding $\hat e_{\texttt{image}}$ and the predicted text embedding $\hat e_{\texttt{text}}$.
To more efficiently utilize the 1-image-3-fMRI data in the NSD \cite{allen2022massive} and alleviate the training-inference data distribution gap, we design a positive sample mixup augmentation.

\paragraph{fMRI Vision Tokens.}
Once the tokenizer is trained, we can obtain the fMRI token $x^{\texttt{L}}, x^{\texttt{R}}$ as a representation of the raw fMRI data.
However, in our visual reconstruction task, only the visual brain regions are meaningful. 
To eliminate redundant fMRI information, we select only the tokens corresponding to the visual brain regions, \textit{i.e.}, $x^\texttt{L}_{\texttt{vis}}, x^\texttt{R}_{\texttt{vis}}$.
To isolate visually responsive regions, we apply the ``NSDGeneral'' ROI mask from NSD~\cite{allen2022massive}, selecting only those fMRI tokens corresponding to these visual areas.
Then, we concatenate the two hemisphere tokens to obtain the tokens that are input to the fMRI encoder: $x=\texttt{concat}(x^\texttt{L}_{\texttt{vis}}, x^\texttt{R}_{\texttt{vis}})$.

\paragraph{Positive Sample Mixup.}
In NSD \cite{allen2022massive}, each image is paired with three fMRI scans, \textit{i.e.}, in embedding space, $e_{\texttt{image}}, e_{\texttt{text}}\Leftrightarrow  x_1, x_2, x_3$.
Here, $x_1, x_2, x_3$ represent the fMRI vision tokens from the three scans, and ``$\Leftrightarrow$'' means ``corresponds to''.

Previous works use $e_{\texttt{image}}, e_{\texttt{text}} \Leftrightarrow x_j$ for training and $e_{\texttt{image}}, e_{\texttt{text}} \Leftrightarrow \texttt{mean}(x_1, x_2, x_3)$ for inference, creating a mismatch between training and testing distributions
To resolve this, we propose positive sample mixup, which randomly combines the three fMRI scans for training, reflecting the eventual inference step more accurately. Specifically, we form a mixup sample:
\begin{align}
    \tilde x_k=\alpha_k x_1+\beta_kx_2+\gamma_kx_3.
\end{align}
Here, $\alpha_k$, $\beta_k$ and $\gamma_k$ are random numbers uniformly sampled from the convex polytope that satisfies the linear constraints defined by equation (\ref{eq:alpha_k_beta_k}).
\begin{align}
    \label{eq:alpha_k_beta_k}
    \begin{cases}
    0 \leq \alpha_k, \beta_k, \gamma_k \leq 1 \\
    \alpha_k+\beta_k+\gamma_k=1
    \end{cases}
\end{align}
Let the number of mixup samples be $K$. 
Then, for a given image, we obtain $K$ corresponding fMRI augmentation data samples: $e_{\texttt{image}}, e_{\texttt{text}}\Leftrightarrow \tilde x_k$ ($k\in[1:K]$).
During contrastive learning, these $K$ samples are used as positive samples with respect to each other (refer to the following “Learning Objectives” part).
In inference time, we set $K=1$, $\alpha_1=\frac{1}{3}$ and $\beta_1=\frac{1}{3}$, which is equivalent to $\tilde x_1=\texttt{mean}(x_1, x_2, x_3)$.
Our positive sample mixup efficiently leverages the data while mitigating the data distribution gap between training and inference.
In practice, we apply positive sample augmentation to a $\lambda$ proportion of the data within a single batch.

Note. For simplicity, we use the vision tokens $x_1, x_2, x_3$ to illustrate mixup. In practice, we apply mixup at the tokenizer’s input level, \textit{e.g.}, $F_1^{\texttt{L}}, F_2^{\texttt{L}}, F_3^{\texttt{L}}$, to avoid confusion and ensure direct consistency with the rest of our pipeline.

\begin{figure}[t]
    \centering
    \includegraphics[width=0.42\textwidth]{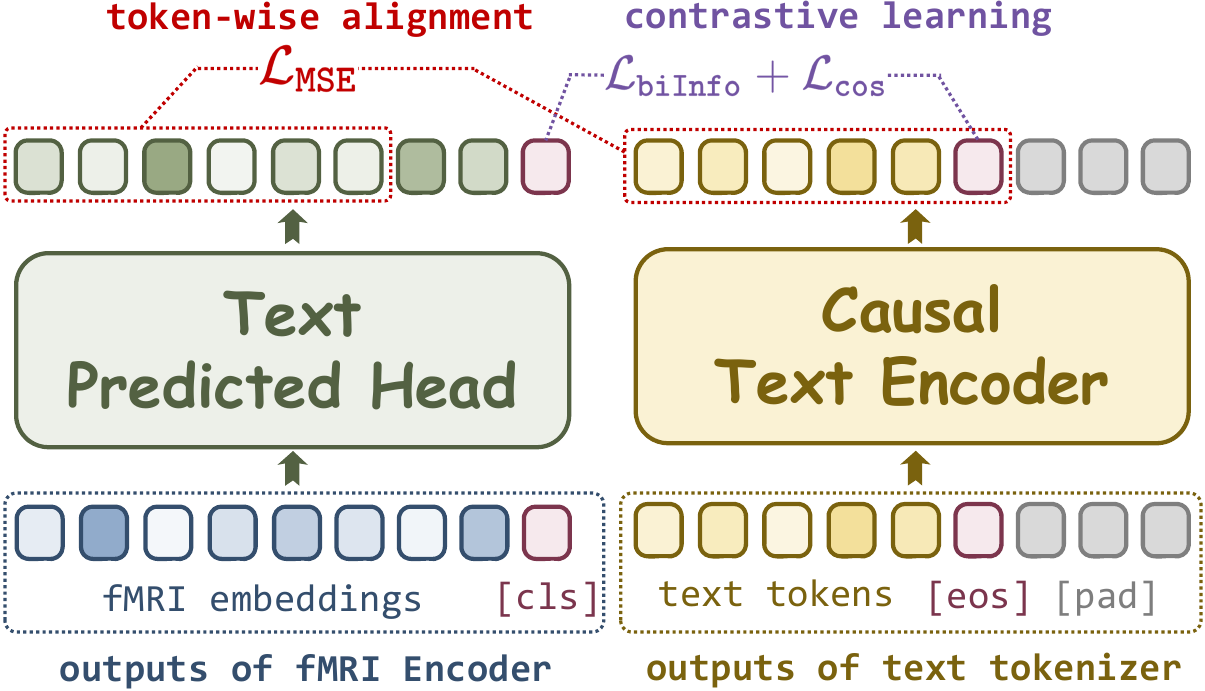}
    \caption{
    In fMRI-text alignment, we only compute the loss for tokens that carry semantic information, unlike previous work that simply aligns all tokens.
    }
    \label{fig:alignment}
\end{figure}

\paragraph{Contrastive Learning.}
Assume that for a training sample $e_{\texttt{image}}, e_{\texttt{text}}\Leftrightarrow x_k$, the prediction of the fMRI encoder is: $\hat e_{\texttt{image}, k}, \hat e_{\texttt{text}, k}=\mathcal E_{\texttt{fMRI}}(x_k)$.
We compute the bidirectional InfoNCE loss \cite{oord2018representation} for both the image and the text separately.
Taking the image as an example, the bidirectional InfoNCE for sample $\hat e_{\texttt{image}, k}$ is defined as:
\begin{align}
\begin{aligned}
    \mathcal L_{\texttt{biInfo}}^{\texttt{image}}=&-\log\frac{K\cdot \exp(\hat e_{\texttt{image}, k}\cdot e_{\texttt{image}}/\tau)}{\sum\limits_{\texttt{img}=1}^bK\cdot\exp(\hat e_{\texttt{image}, k}\cdot e_{\texttt{img}}/\tau)}\\
    &-\log\frac{\sum\limits_{k}^K \exp (e_{\texttt{image}}\cdot \hat e_{\texttt{image}, k}/\tau)}{\sum\limits_{\texttt{img}=1}^b\sum\limits_{k}^K\exp(e_{\texttt{image}}\cdot e_{\texttt{img}, k}/\tau)}.
\end{aligned}
\end{align}
Here, $b$ represents the batch size, $\tau$ is the temperature hyperparameter, and $K$ is the number of augmentation samples (where $\texttt{image}$ is augmented with probability $\lambda$) or 3 (when $\texttt{image}$ is not augmented with probability $1-\lambda$).
The bidirectional InfoNCE for the text $\mathcal L_{\texttt{biInfo}}^{\texttt{text}}$ is defined in the same way.
The final bidirectional InfoNCE with positive sample augmentation is: $\mathcal L_{\texttt{biInfo}}=\mathcal L_{\texttt{biInfo}}^{\texttt{image}}+\mathcal L_{\texttt{biInfo}}^{\texttt{text}}$.
To further reduce the distance between embeddings, we also use a cosine similarity loss:
\begin{align}
    \mathcal L_{\texttt{cos}}^{\texttt{image}}=1-\frac{\hat e_{\texttt{image}, k}\cdot e_{\texttt{image}, k}}{||\hat e_{\texttt{image}, k}||\cdot ||e_{\texttt{image}, k}||}.
\end{align}
%

\paragraph{Token-Wise Alignment.}
Previous work \cite{quan2024psychometry, wang2024mindbridge, huo2025neuropictor} have shown that token-wise alignment is more effective than performing contrastive learning only on the \texttt{[cls]} token, \textit{i.e.}, by computing the prediction loss for each token.
We adopt this approach for text alignment and further refine it to better suit the textual domain.
Because the CLIP \cite{radford2021learning} text encoder is causal, any tokens generated after the \texttt{[eos]} marker in its output embeddings no longer hold meaningful information.
As shown in Fig. \ref{fig:alignment}, we only compute the loss for the parts with semantic information in token-wise alignment.
For the image, all tokens participate in the loss calculation.
The loss for token-wise alignment is MSE ($\mathcal L_{\texttt{MSE}}$).
Finally, the total loss function is $\mathcal L=\mathcal L_{\texttt{biInfo}}+\mathcal L_{\texttt{cos}}+\mathcal L_{\texttt{MSE}}$.

\subsection{Image Reconstruction}
\label{sec:method-image_reconstruction}
After obtaining the reconstructed CLIP space semantic embeddings $\hat e_{\texttt{image}}$ and $\hat e_{\texttt{text}}$, we need an image decoder to reconstruct the image.
We choose versatile diffusion \cite{xu2023versatile} as our image decoder, a multimodal diffusion model guided by both CLIP image embeddings and CLIP text embeddings.
During inference, we also use classifier-free guidance \cite{rombach2022high} and Ecphory Mechanism introduced by Psychometry \cite{quan2024psychometry}.

\section{Experiments}
\begin{figure}[t]
    \centering
    \includegraphics[width=0.45\textwidth]{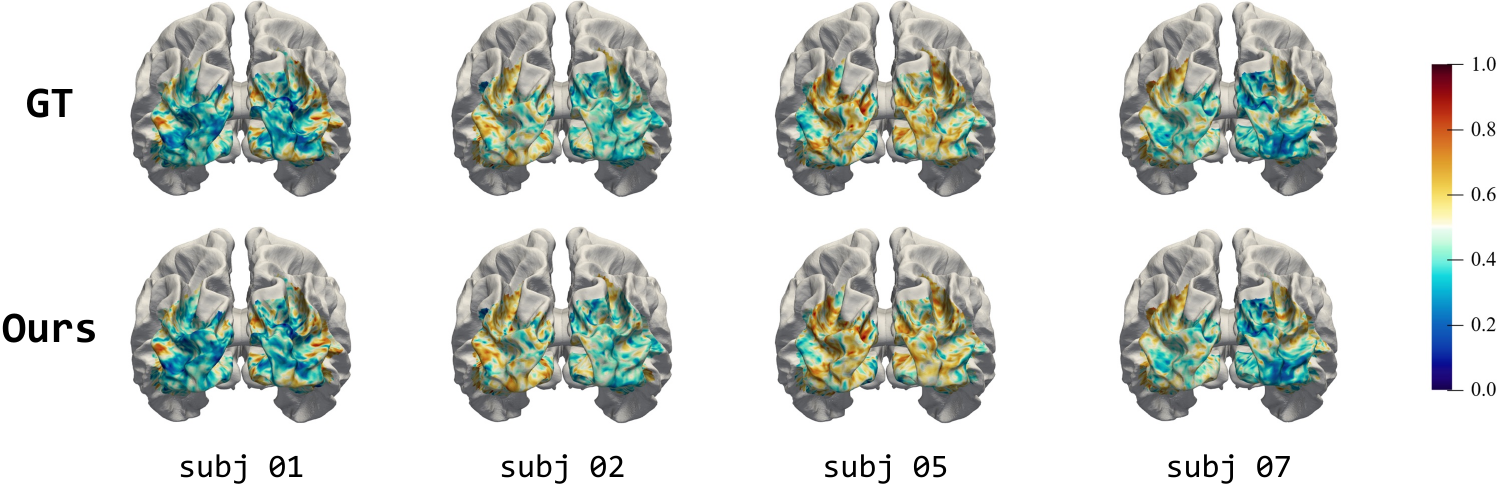}
    \caption{
    fMRI vision voxels reconstruction results using the sphere tokenizer on the NSD \cite{allen2022massive} \texttt{test}.
    The sphere tokenizer demonstrates excellent encoding-decoding capabilities across the four subjects.
    See related analysis in \textsection \ref{sec:experiments-sphere_tokenizer}.
    }
    \label{img:fmri_rec}
\end{figure}

\paragraph{Evaluation Metrics.}
We follow the standard settings in the previous work \cite{scotti2024reconstructing, quan2024psychometry, wang2024mindbridge, shen2025neuro} and use eight image quality evaluation metrics.
For low-level properties, we apply pixel-wise correlation (\texttt{PixCorr}), structural similarity (\texttt{SSIM}) \cite{wang2004image}, \texttt{AlexNet}(\texttt{2}), and \texttt{AlexNet}(\texttt{5}) \cite{krizhevsky2012imagenet}.
For high-level evaluation, we adapt \texttt{Inception} \cite{szegedy2016rethinking}, \texttt{CLIP} \cite{radford2021learning}, \texttt{EffNet-B} \cite{tan2019efficientnet}, and \texttt{SwAV} \cite{caron2020unsupervised}.
The reported results are the averages computed across four subject.

\paragraph{Reproducibility.}
The model is trained on one NVIDIA A800 GPU.
Inference is conducted on the same machine.
Other hyperparameters are at Appendix \textsection \ref{appendix:training_details}.

\subsection{Sphere Tokenizer}
\label{sec:experiments-sphere_tokenizer}

The proposed sphere tokenizer maps the raw fMRI data into input embeddings for the encoder, and it is trained using an autoencoder architecture.
Since the sphere tokenizer effectively handles data compression before the fMRI encoder, we are interested in its ability to reconstruct fMRI data.
We present the fMRI reconstruction results of visual voxels on the NSD \texttt{test} using the sphere tokenizer autoencoder (Fig. \ref{img:fmri_rec}).
The displayed data has been min-max normalized to range [0:1] across all visual voxels of each sample.
The qualitative results demonstrate a high similarity between the reconstructed fMRI and the groundtruth.
This shows that the sphere tokenizer can retain a large amount of information from the visual voxels.
We also provide more evaluation results for the sphere tokenizer in Appendix \textsection \ref{appendix:more_results}.

\subsection{Comparison to Previous Work}
\label{sec:experiments_comparison}

\begin{table*}
    \centering
    \resizebox{\linewidth}{!}{
    \begin{tabular}{rccccccccc}
    
        \specialrule{1.5pt}{2pt}{2pt}
        
        \multirow{2.5}{*}{Method} & 
        \multirow{2.5}{*}{\makecell{\# fMRI \\ Vision Voxels}} &
        \multicolumn{4}{c}{Low-Level} &
        \multicolumn{4}{c}{High-Level}
        \\ 
        \cmidrule(lr){3-6} \cmidrule(lr){7-10} 
        & &
        \texttt{PixCorr} $\uparrow$ &
        \texttt{SSIM} $\uparrow$ &
        \texttt{AlexNet}(\texttt{2}) $\uparrow$ &
        \texttt{AlexNet}(\texttt{5}) $\uparrow$ &
        \texttt{Inception} $\uparrow$ &
        \texttt{CLIP} $\uparrow$ &
        \texttt{EffNet-B} $\downarrow$ &
        \texttt{SwAV} $\downarrow$ \\
        
        \specialrule{0.5pt}{2pt}{2pt}
        
        Mind-Vis \cite{chen2023seeing} & 
        14K-\texttt{1D} & 
        0.067 & 
        0.196 &
        67.7\% &
        74.2\% &
        67.9\% &
        69.3\% &
        0.898 &
        0.513 \\
        
       MindEye \cite{scotti2024reconstructing} &
       14K-\texttt{1D} &
       0.129 &
       0.255 &
       84.2\% &
       89.2\% &
       84.1\% &
       85.0\% &
       0.812 &
       0.487 \\

       Neuro-Vision \cite{shen2025neuro} &
       14K-\texttt{1D} &
       0.265 &
       0.357 &
       93.1\% &
       97.1\% &
       96.8\% &
       97.5\% &
       0.633 &
       0.321 \\

       Psychometry \cite{quan2024psychometry} &
       14K-\texttt{1D} &
       0.297 &
       0.340 &
       96.4\% &
       98.6\% &
       95.8\% &
       96.8\% &
       0.628 &
       0.345 
       \\

       UMBRAE \cite{xia2025umbrae} &
       14K-\texttt{1D} &
       0.283 &
       0.328 &
       93.9\% &
       96.7\% &
       93.4\% &
       94.1\% &
       0.700 &
       0.393 \\

       NeuroPictor \cite{huo2025neuropictor} &
       14K-\texttt{1D} &
       0.141 &
       0.349 &
       91.4\% &
       95.7\% &
       88.3\% &
       88.9\% &
       0.722 &
       0.417 \\

       MindBridge \cite{wang2024mindbridge} &
       14K-\texttt{1D} &
       0.151 &
       0.263 &
       87.7\% &
       95.5\% &
       92.4\% &
       94.7\% &
       0.712 &
       0.418 \\

       \specialrule{0.5pt}{2pt}{2pt}

       Gu \textit{et al}. \cite{gu2024decoding} &
       7K-\texttt{sphere} &
       0.103 &
       0.264 &
       - &
       - &
       - &
       - &
       0.892 &
       0.508 \\

       MindBridge \cite{wang2024mindbridge} &
       9K-\texttt{1D} &
       0.143 &
       0.307 &
       75.0\% &
       86.7\% &
       83.8\% &
       87.0\% &
       0.769 &
       0.413 \\
       
       Ours &
       9K-\texttt{sphere} &
       0.165 &
       0.305 &
       78.2\% &
       89.0\% &
       85.1\% &
       88.3\% &
       0.733 &
       0.398 \\
       
       \specialrule{1.5pt}{2pt}{2pt}
        
    \end{tabular}
    }
    \caption{
    Quantitative comparison results on NSD \cite{allen2022massive} \texttt{test}.
    All methods are trained using a single model on 4 subjects.
    Our method uses a lower fMRI resolution (9K) compared to SOTA (14K) due to the standard settings of SphericalUnet \cite{zhao2019spherical}.
    See related analysis in \textsection \ref{sec:experiments_comparison}.
    }
    \label{tab:compare}
    
\end{table*}

\begin{figure*}[t]
    \centering
    \includegraphics[width=0.8\textwidth]{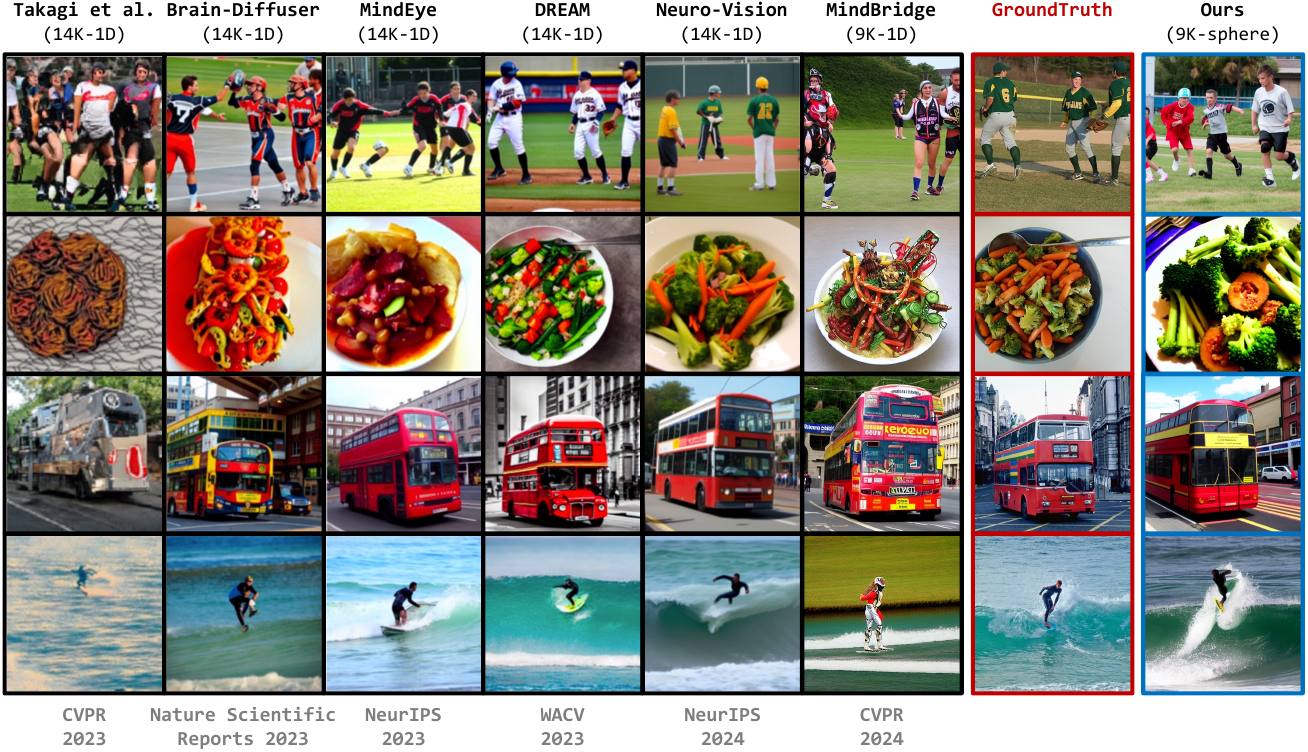}
    \caption{
    Qualitative results on NSD \cite{allen2022massive} \texttt{test} subj01.
    In comparison to previous single-subject models \cite{takagi2023high, ozcelik2023natural, scotti2024reconstructing, xia2024dream} and recent cross-subject models \cite{shen2025neuro, wang2024mindbridge}, our model is trained on lower fMRI resolution without subject-specific parameters. 
    The visual results of reconstruction demonstrate the effectiveness of our approach.
    See related discussion in \textsection \ref{sec:experiments_comparison}.
    }
    \label{fig:results-image}
\end{figure*}

\paragraph{Quantitative Results.}
We compare our model with eight previous methods: Mind-Vis \cite{chen2023seeing}, MindEye \cite{scotti2024reconstructing}, Gu \textit{et al}. \cite{gu2024decoding}, Psychometry \cite{quan2024psychometry}, UMBRAE \cite{xia2025umbrae}, NeuroPictor \cite{huo2025neuropictor}, Neuro-Vision \cite{shen2025neuro} and MindBridge \cite{wang2024mindbridge}.
The quantitative evaluation results are shown in Tab. \ref{tab:compare}.
All models are trained using a single model in the 4 subjects,  and the results are the average performance in the 4 subjects.
Due to the fact that the sphere tokenizer is based on SphericalUnet \cite{zhao2019spherical}, which only supports standard spheres, the fMRI resolution we used is different from other studies.
We present the fMRI resolution of each model in Tab. \ref{tab:compare}, expressed in terms of the number of vision voxels.
``14K-\texttt{1D}'' indicates that each subject has approximately 14K voxels\footnote{subj01: 15,247, subj02: 14,278, subj05: 13,039, and subj07: 12,682.} and treats them as a one-dimensional signal.
Gu \textit{et al.} applies the spherical convolution framework on the FreeSurfer 32k surface, containing 7,531 visual voxels (7K-\texttt{sphere}). 
The data we used (9K-\texttt{sphere}) contains 9,548 visual voxels, which entails lower computational costs compared to the 14K data (\textit{e.g.}, in contrast to 7T, our model is also compatible with 3T fMRI data).
Compared to methods that also use a spherical framework as the fMRI encoder (Gu \textit{et al}. \cite{gu2024decoding}), our model shows a significant performance improvement (\textit{e.g.}, $0.103/0.264\to0.165/0.305$ in low-level metrics \texttt{PixCorr} and \texttt{SSIM}, and $0.892/0.508\to0.733/0.398$ in high-level metrics \texttt{EffNet-B} and \texttt{SwAV}).
However, compared to the current state-of-the-art, our model performs relatively worse due to the use of a lower fMRI resolution.
To demonstrate that this performance decline is solely due to resolution (14K \textit{vs.} 9K), we replicate MindBridge \cite{wang2024mindbridge} at the same resolution (9K-\texttt{1D}).
At the same resolution of fMRI data, our model improves upon MindBridge by $0.022/$$-$$0.002/3.2\%/2.3\%$ on the four low-level metrics and by $1.3\%/1.3\%/0.036/0.015$ on the four high-level metrics.
Additionally, our model outperforms higher resolution models in certain aspects, \textit{e.g.}, our model shows improvements over MindEye \cite{scotti2024reconstructing} by $1.0\%/3.3\%/0.079/0.089$ on the four high-level metrics.

\begin{figure}[t]
    \centering
    \includegraphics[width=0.45\textwidth]{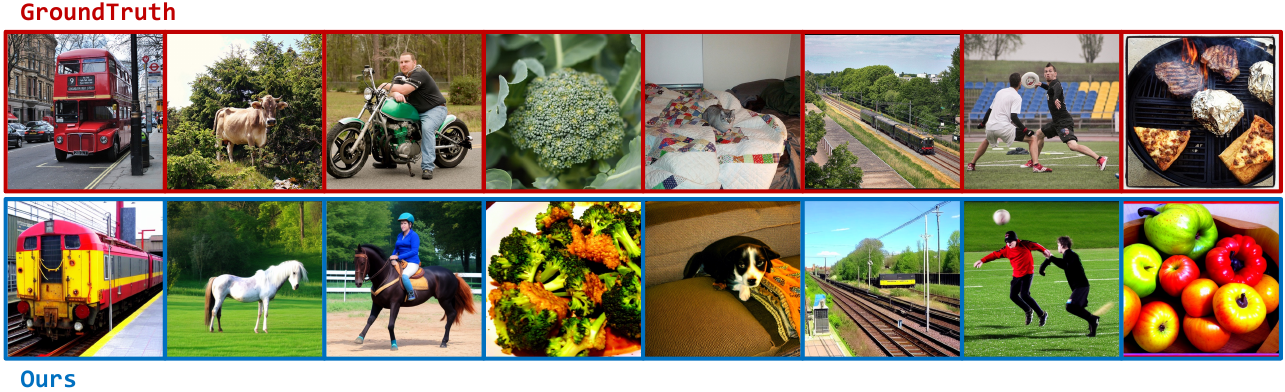}
    \caption{
    Failed cases in the NSD \cite{allen2022massive} \texttt{test} subj01.
    In these cases, although the reconstruction results are not entirely satisfactory, our model has managed to reconstruct partial or analogous semantics.
    See related analysis in \textsection \ref{sec:experiments_comparison}.
    }
    \label{fig:results-image_bad_case}
\end{figure}

\begin{table*}
    \centering
    \resizebox{\linewidth}{!}{
    \begin{tabular}{clcccccccc}
    
        \specialrule{1.5pt}{2pt}{2pt}

        \multirow{2.5}{*}{\#} &
        \multirow{2.5}{*}{Setting} & 
        \multicolumn{4}{c}{Low-Level} &
        \multicolumn{4}{c}{High-Level}
        \\ 
        \cmidrule(lr){3-6} \cmidrule(lr){7-10} & &
        \texttt{PixCorr} $\uparrow$ &
        \texttt{SSIM} $\uparrow$ &
        \texttt{AlexNet}(\texttt{2}) $\uparrow$ &
        \texttt{AlexNet}(\texttt{5}) $\uparrow$ &
        \texttt{Inception} $\uparrow$ &
        \texttt{CLIP} $\uparrow$ &
        \texttt{EffNet-B} $\downarrow$ &
        \texttt{SwAV} $\downarrow$ \\
        
        \specialrule{0.5pt}{2pt}{2pt}

       1 &
       Conv1$\times$1 &
       0.190 &
       0.319 &
       64.0\% &
       78.9\% &
       76.7\% &
       79.4\% &
       0.819 &
       0.405 \\

       2 &
       ROI Embedder &
       0.139 &
       0.312 &
       73.6\% &
       85.4\% &
       83.4\% &
       85.5\% &
       0.771 &
       0.415 \\

       \specialrule{0.5pt}{2pt}{2pt}

       3 &
       w/o structure condition  &
       0.146 &
       0.311 &
       75.4\% &
       85.9\% &
       83.3\% &
       86.0\% &
       0.758 &
       0.410 \\

       4 &
       w/o positional condition &
       0.142 &
       0.302 &
       74.0\% &
       85.0\% &
       84.3\% &
       86.0\% &
       0.758 &
       0.413 \\

       \specialrule{0.5pt}{2pt}{2pt}

       5 &
       w/o token-wise $\mathcal L_{\texttt{MSE}}$ &
       0.118 &
       0.301 &
       69.6\% &
       79.8\% &
       79.2\% &
       81.6\% &
       0.809 &
       0.448 \\

       6 &
       full tokens $\mathcal L_{\texttt{MSE}}^{\texttt{text}}$ &
       0.152 &
       0.312 &
       75.6\% &
       86.7\% &
       83.9\% &
       86.7\% &
       0.751 &
       0.408 \\

       \specialrule{0.5pt}{2pt}{2pt}
       
       7 &
       Full Model &
       0.165 &
       0.305 &
       78.2\% &
       89.0\% &
       85.1\% &
       88.3\% &
       0.733 &
       0.398 \\
       
       \specialrule{1.5pt}{2pt}{2pt}
        
    \end{tabular}
    }
    \caption{
    Ablation study (\textsection \ref{sec:experiments-ablation}) on sphere tokenizer on NSD \cite{allen2022massive} \texttt{test}.
    The experimental results substantiate the efficacy of the sphere tokenizer, the structural and positional conditions, as well as the token-wise alignment.
    }
    \label{tab:ablation}
    
\end{table*}

\begin{figure}[t]
    \centering
    \includegraphics[width=0.45\textwidth]{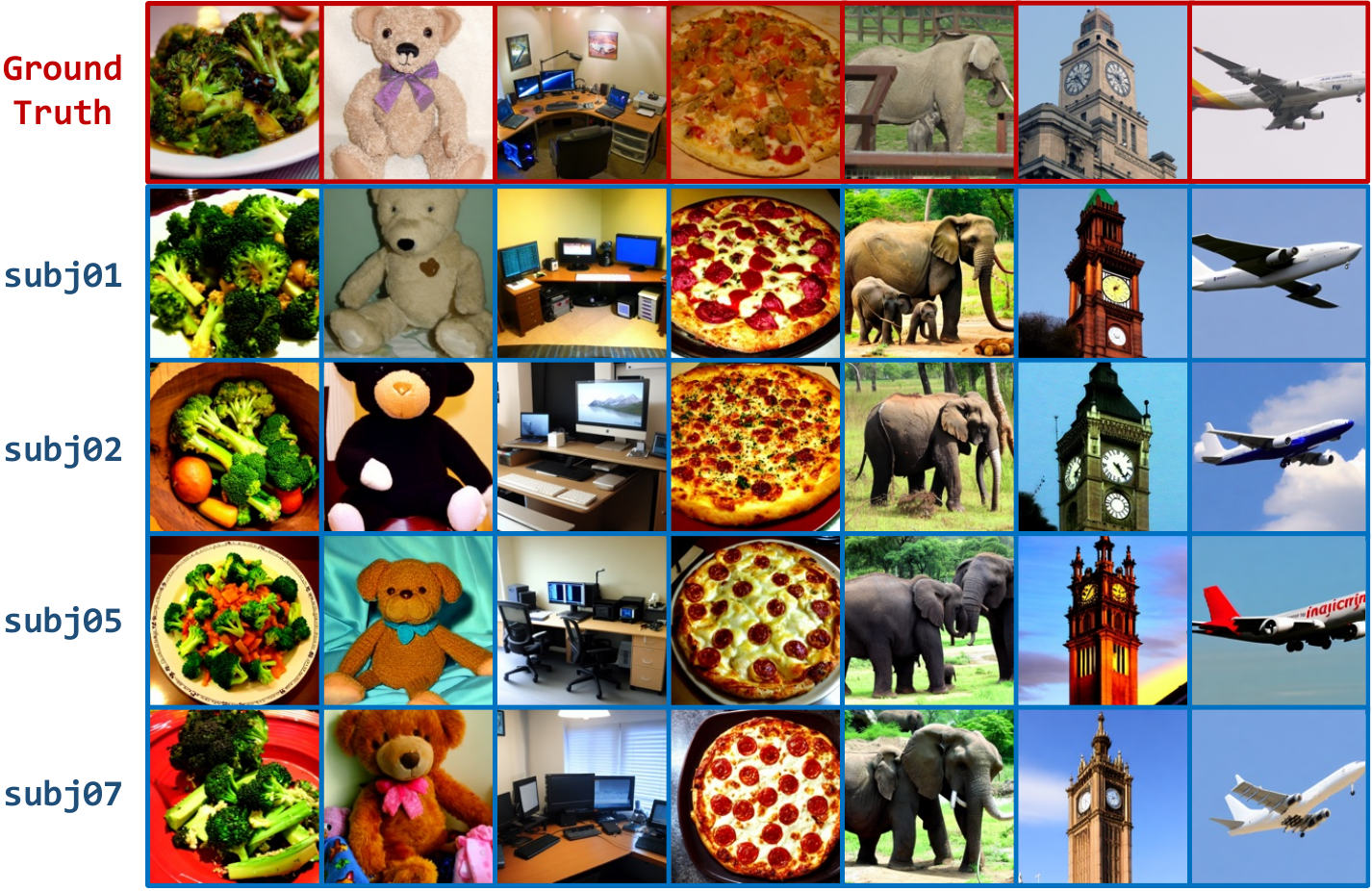}
    \caption{
    Visual samples on NSD \cite{allen2022massive} \texttt{test} over 4 subjects.
    The results demonstrate the cross-subject robustness of our model.
    }
    \label{fig:results-image_4_subjs}
\end{figure}

\paragraph{Qualitative Results.}
We present some brain decoding samples on the NSD \cite{allen2022massive} \texttt{test} for subj01 in Fig \ref{fig:results-image}.
The baseline models are Takagi \textit{et al.} \cite{takagi2023high}, Brain-Diffuser \cite{ozcelik2023natural}, MindEye \cite{scotti2024reconstructing}, DREAM \cite{xia2024dream}, Neuro-Vision \cite{shen2025neuro} and MindBridge \cite{wang2024mindbridge}.
The results of MindBridge are obtained on low fMRI resolution (9K-\texttt{1D}).
The image quality reconstructed by our method is not worse than that of other methods.
We also provide the failed cases in Fig. \ref{fig:results-image_bad_case}.
In these cases, our model captures partial but incomplete semantics (\textit{e.g.}, riding a motorcycle $\to$  riding a horse, cat $\to$ dog, and frisbee $\to$ soccer), which affects quantitative metrics.
We also present the reconstruction results for four subjects in Fig. \ref{fig:results-image_4_subjs}, demonstrating the model’s cross-subject robustness.
For more visualization results, please refer to Appendix \textsection \ref{appendix:more_results}.

\subsection{Ablation Study}
\label{sec:experiments-ablation}

\paragraph{Different Tokenizers.}
We first evaluate the effectiveness of the proposed sphere tokenizer.
We consider two other variants of the fMRI tokenizer: Conv1$\times$1 \cite{chen2023seeing} and ROI Embedder \cite{qian2023joint, quan2024psychometry}.
The results are shown in Tab. \ref{tab:ablation}.
The results show that Conv1$\times$1 has poor ability to capture the spatial and structural properties of fMRI data.
When replacing the fMRI tokenizer from a Conv1$\times$1 with ROI Embedder that considers the spatial characteristics of different brain regions, we observe a significant performance improvement.
When the modeling of fMRI spatial information is upgraded from the ROI level (ROI Embedder, \#2) to the voxel level (ours, Sphere Tokenizer, \#7), we also observe performance improvement (\textit{e.g.}, the performance increase $4.6\%/3.6\%/2.8\%/2.8\%$ across \texttt{AlexNet}(\texttt{2}), \texttt{AlexNet}(\texttt{5}), \texttt{Inception} and \texttt{CLIP}).

\paragraph{Cortex Structure and Positional Embedding.}
Tab. \ref{tab:ablation} also investigates the impact of cortex structure and positional embedding in the sphere tokenizer.
As the results show, the introduction of  the structural information of the cerebral cortex into sphere tokenizer brings a performance boost (\textit{e.g.}, $83.4\%/85.5\%/0.771/0.415$ $\to$ $85.1\%/88.3\%/0.733/0.398$ on all high-level metrics from \#3 to \#7).
This suggests that subject-specific brain structural information facilitates cross-subject brain decoding.
Additionally, when removing the sphere positional embedding (\#4), we observe a decline in performance.

\paragraph{Token-Wise Alignment.}
We first remove the token-wise alignment, optimizing only the \texttt{[cls]} token, which led to a performance drop (\#5). 
Additionally, we observe that the attention weights for each token are almost identical, causing the transformer to degrade into an MLP.
When optimizing all tokens during the fMRI-text alignment (\#6), the meaningless \texttt{[pad]} tokens in the text also contribute to the loss, leading to a certain degree of performance degradation.

\begin{figure}[t]
    \centering
    \includegraphics[width=0.4\textwidth]{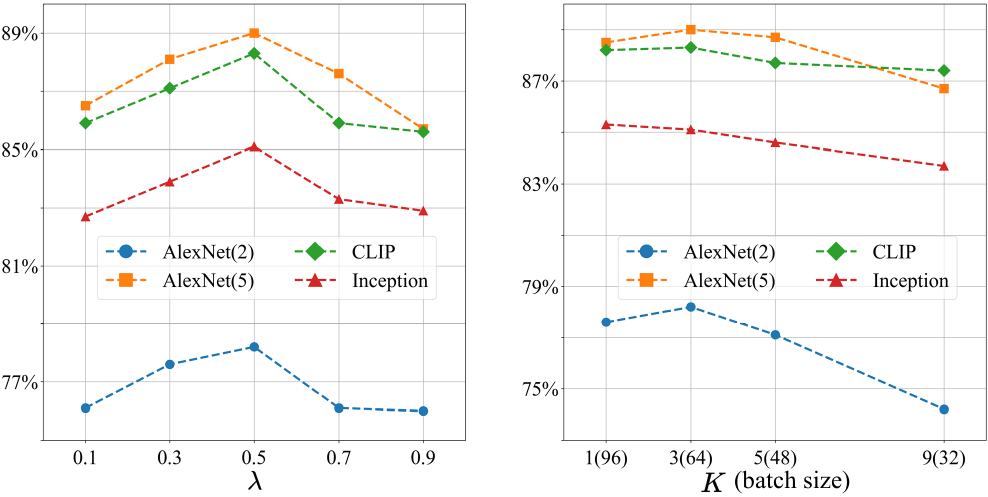}
    \caption{
    The impact of $\lambda$ and $K$ in positive sample mixup (\textsection \ref{sec:experiments-ablation}).
    %
    }
    \label{fig:results-lambda_k}
\end{figure}

\paragraph{Positive Sample Mixup.}
%
We also design ablation experiments to validate the effectiveness of the positive sample mixup.
There are two adjustable hyperparameters here: the mixup ratio $\lambda$ and target mixup samples number $K$.
However, variations in $\lambda$ and $K$ may affect the batch size.
To ensure a fair comparison, we control the expected batch size (or number of images in a batch) to be a constant $B$ in the experiments.
%
%
Let the batch size in an experiment be $b$, then the following condition holds: $\lambda Kb+3(1-\lambda)b=3B$.
Then we have $b=\frac{3B}{3+(K-3)\lambda}$.
We show the impact of $\lambda$ and $K$ in Fig. \ref{fig:results-lambda_k}.
Both $\lambda$ and $K$ have a peak effect on performance, reflecting the trade-off in positive sample mixup.
When $\lambda$ is too low, the augmentation strength is insufficient. 
On the other hand, when $\lambda$ is too high, almost all data become synthetic, affecting the characteristics of the original data.
For $K$, a value that is too small makes the augmentation too weak, while a value that is too large leads to an insufficient number of negative samples within a batch.
In practice, our base batch size is $B = 64$. 
To ensure that $b$ is an integer, we conduct four sets of experiments on $K$ with the optimal $\lambda=0.5$: $K=1,3,5,9$ with $b=96,64,48,32$.
Based on the results, we ultimately choose $\lambda = 0.5$ and $K = 3$.

\subsection{Biological Interpretability}
\label{sec:experiments-bio_interpretability}

\begin{figure}[t]
    \centering
    \includegraphics[width=0.4\textwidth]{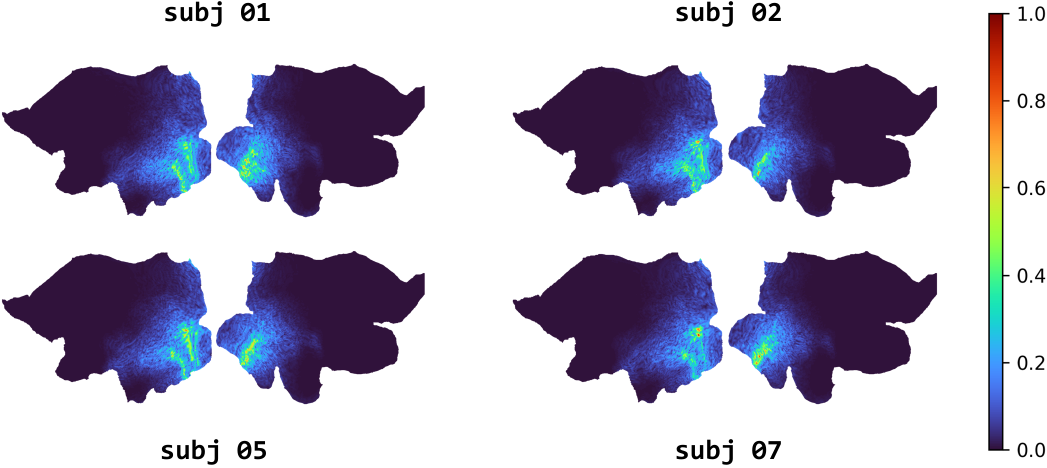}
    \caption{
    Average neural activity representation differential heatmaps on the NSD \cite{allen2022massive} \texttt{test}.
    See analysis in \textsection \ref{sec:experiments-bio_interpretability}.
    }
    \label{fig:results-heatmap}
\end{figure}

To further our understanding of the importance of different brain voxels in the vision brain decoding task, we compute the average fMRI heatmaps using Grad-CAM \cite{selvaraju2017grad} across all samples in the NSD \cite{allen2022massive} \texttt{test} (Fig. \ref{fig:results-heatmap}).

\paragraph{Brain Region.}
The bright spots in are distributed across the visual regions, with high-level areas being more important than low-level areas.
Our model focuses more on high-level areas due to the alignment with CLIP \cite{radford2021learning} semantics.

\paragraph{Cross Subjects.}
The model explores subject-independent patterns and features when handling cross subject fMRI data.
As shown in Fig. \ref{fig:results-heatmap}, the four subjects have very similar fMRI heatmaps.
This explains why our model is able to achieve cross-subject vision brain decoding.

\section{Discussion and Conclusion}
We approach fMRI-image reconstruction from a novel perspective.
We introduce the sphere tokenizer, which treats fMRI signals as 2D data on spherical surface instead of 1D.
Considering the individual brain structure-function relationships and differences from cortical locations, we additionally incorporate structure and positional condition for the sphere tokenizer.
Furthermore, we use positional embedding to help the model better understand the function of different brain regions.
In addition, we design positive sample mixup to more efficiently utilize training data and alleviate the data distribution gap between training and inference.
Our model achieves satisfactory performance even under the constraint of fMRI resolution.

\section{Acknowledgements}
This work is funded by Guangdong Provincial Key Laboratory of Human Digital Twin (2022B1212010004),
General Program of the National Natural Science Foundation of China (NSFC-62471185), Guangdong Basic and Applied Basic Research Foundation (2024A1515010180), and Key-Area Research and Development Program of Guangdong Province (2023B0303040001).

{
    \small
    \bibliographystyle{ieeenat_fullname}
    \bibliography{reference}
}

\newpage
\maketitlesupplementary
\appendix
\makeatletter
\renewcommand{\thefigure}{A-\arabic{figure}}
\renewcommand{\thetable}{A-\arabic{table}}
\renewcommand{\theequation}{A-\arabic{equation}}
\renewcommand{\thealgocf}{A-\arabic{algocf}}
\makeatother
\setcounter{figure}{0}
\setcounter{table}{0}
\setcounter{equation}{0}
\setcounter{algocf}{0}

\section{Data Preprocessing}
\label{appendix:data_preprocess}
\paragraph{fMRI Data.}
This paper is based on the Natural Scenes Dataset (NSD)\footnote{\hyperlink{https://naturalscenesdataset.org}{\texttt{https://naturalscenesdataset.org}}} \cite{allen2022massive}.
Since we use spherical data as model's input, we employ the data preprocessed with FreeSurfer, which is provided by the official NSD dataset.
Taking subj01 as an example, the original fMRI path we use is:
\begin{align*}
&\texttt{nsddata\_betas/ppdata/subj01/fsaverage} \\
&~~\texttt{/betas\_fithrf\_GLMdenoise\_RR/}
\end{align*}
The data values here are single-trial beta weights estimated by applying a general linear model (GLM) to the raw fMRI time series, representing the voxel-wise response and its correlation with visual stimuli.
The data provided by the official source are registered to the FreeSurfer standard fsaverage7 surface.
We resample the data to a 40,962 sphere to match the compatibility with SphericalUNet \cite{zhao2019spherical}.
Resampling is performed using the open-source SphericalUNet \cite{zhao2019spherical} package.
Then, we independently apply zero-score normalization for each voxel within the \texttt{train} data of a subject.
The \texttt{val} and \texttt{test} data are zero-centered using the mean and variance from the \texttt{train} data.
The data split follows the standard setup used in previous works.

\paragraph{Cortex Structure Data.}
Still taking subj01 as an example, our cortical structural data comes from the path:
$$
\texttt{nsddata/freesurfer/subj01/surf}
$$
We use four types of structural information: cortical thickness, surface area, sulcal depth, and curvature.
Taking the left hemisphere as an example, we use the following four files: \texttt{lh.thickness}, \texttt{lh.area}, \texttt{lh.sulc}, and \texttt{lh.curv}.
We apply the same method as with the fMRI data to resample it to the fsaverage6 surface.
Then, we performed zero-score normalization on each individual file (\textit{i.e.}, each hemisphere of each subject).

\paragraph{Image-Text Pair.}
All images in the NSD dataset are derived from the MS-COCO \cite{lin2014microsoft} dataset.
The text annotations we use are from the official MS-COCO\footnote{\hyperlink{http://images.cocodataset.org}{\texttt{http://images.cocodataset.org}}} dataset.
\makeatletter
\renewcommand{\thefigure}{B-\arabic{figure}}
\renewcommand{\thetable}{B-\arabic{table}}
\renewcommand{\theequation}{B-\arabic{equation}}
\renewcommand{\thealgocf}{B-\arabic{algocf}}
\makeatother
\setcounter{figure}{0}
\setcounter{table}{0}
\setcounter{equation}{0}
\setcounter{algocf}{0}

\section{Technical Details}

\paragraph{Random Rotation of Sphere Positional Emb.}
The pseudocode is shown in Algorithm \ref{algo_appendix:rotation}.
In the algorithm, we apply Rodrigues’ rotation formula, a mathematical tool used for rotating 3D coordinates in 3D Euclidean space.
\IncMargin{1em}
\begin{algorithm} 
\SetKwData{Left}{left}
\SetKwData{This}{this}
\SetKwData{Up}{up} 
\SetKwFunction{Union}{Union}
\SetKwFunction{FindCompress}{FindCompress} \SetKwInOut{Input}{input}
\SetKwInOut{Output}{output}
	
	\Input{Original spherical coordinates $\mathbf x\in\mathbb R^{n\times3}$ \\ Maximum rotation angle $\theta_{\max}$} 
	\Output{Augmented coordinates $\mathbf x'\in\mathbb R^{n\times3}$}
    
	 \BlankLine 
     \textcolor[rgb]{0.3, 0.7, 0.3}{\# get a random rotation axis $\mathbf v$} \\
     $\phi\sim\mathcal U(0, 2\pi)$ \\
     $\varphi\sim\mathcal U(0,\pi)$ \\
     $\mathbf v=(v_x, v_y, v_z)=(\sin\phi\cos\varphi, \sin\phi\sin\varphi,\cos\phi)$ \\

     \BlankLine 

     \textcolor[rgb]{0.3, 0.7, 0.3}{\# get a random rotation angle $\theta$} \\
     $\theta\sim\mathcal U(0, \theta_{\max})$

     \BlankLine

     \textcolor[rgb]{0.3, 0.7, 0.3}{\# apply Rodrigues’ rotation formula} \\
     $\mathbf K = \begin{pmatrix} 0 & -v_z & v_y \\ v_z & 0 & -v_x \\ -v_y & v_x & 0 \end{pmatrix}$, $\mathbf I = \begin{pmatrix} 1 & 0 & 0 \\ 0 & 1 & 0 \\ 0 & 0 & 1 \end{pmatrix}$ \\
     $\mathbf R=\mathbf I+\sin\theta\cdot\mathbf K+(1-\cos\theta)\cdot\mathbf K^2$\\
     $\mathbf x'=\mathbf x\mathbf R$ \\
     \Return{$\mathbf x'$}

      \caption{Random Rotation Augmentation of Sphere Positional Embedding}
      \label{algo_appendix:rotation} 
 \end{algorithm}
 \DecMargin{1em} 
\paragraph{Positive Sample Mixup.}
We present the pseudocode for positive sample mixup augmentation in Algorithm \ref{algo_appendix:positive_sample_mixup}.
To uniformly sample a point $\mathbf w$ as the mixup weight in the convex polytope defined by equation (\ref{eq:alpha_k_beta_k}), we sample a point from the 3-dimensional Dirichlet distribution $\mathcal D$ with parameters $\mathbf a=(1, 1, 1)$.
The $N$-dimensional Dirichlet distribution is the $N$-dimensional generalization of the Beta distribution, and its probability density function is given by:
\begin{align}
    p_{\mathcal D}(\mathbf w|\mathbf a)= \frac{1}{ B (\mathbf a)}\prod_{n=1}^{N}w_n^{a_n-1}.
    \label{eq_appendix:dist}
\end{align}
Here, $B(n)$ is the normalization factor, ensuring that the probability density function integrates to 1 over the domain:
\begin{align}
    B(\mathbf a)=\frac{\prod\limits_{n=1}^{N}\Gamma(a_n)}{\Gamma\left(\sum\limits_{n=1}^{N}a_n\right)}.
\end{align}

\begin{algorithm} \SetKwData{Left}{left}\SetKwData{This}{this}\SetKwData{Up}{up} \SetKwFunction{Union}{Union}\SetKwFunction{FindCompress}{FindCompress} \SetKwInOut{Input}{input}\SetKwInOut{Output}{output}
	
	\Input{Three fMRI scans $\mathbf x=(x_1, x_2, x_3)$ \\
    Mixup ratio $\lambda$ \\
    Mixup number $K$
    } 
	\Output{Augmented samples $\tilde x_1, \tilde x_2,\cdots$}
    
	 \BlankLine 

    $t\sim\mathcal U(0,1)$\\
    \If{$t<\lambda$}{
        \For{$k\gets1$ \KwTo $K$}{
            $\mathbf a=(1,1,1)$\\
            $\mathbf w\sim\mathcal D(\mathbf a)$ \textcolor[rgb]{0.3, 0.7, 0.3}{~~\# Dirichlet Dist.,~Eq.~(\ref{eq_appendix:dist})}\\
            $\tilde x_k=\mathbf w\cdot \mathbf x$
        }
        \Return{$\tilde x_1, \tilde x_2, \cdots, \tilde x_K$}
    }
    \Else{
        \Return{$x_1, x_2, x_3$}
    }
      \caption{Positive Sample Mixup}
      \label{algo_appendix:positive_sample_mixup} 
 \end{algorithm}

\paragraph{Multi Positive Samples InfoNCE.}
We modify the InfoNCE \cite{oord2018representation} to accommodate the scenario where multiple positive samples exist within a batch.
The algorithm for multi positive samples InfoNCE is in Algorithm \ref{algo_appendix:infonce}.
\begin{algorithm} \SetKwData{Left}{left}\SetKwData{This}{this}\SetKwData{Up}{up} \SetKwFunction{Union}{Union}\SetKwFunction{FindCompress}{FindCompress} \SetKwInOut{Input}{input}\SetKwInOut{Output}{output}
	
\Input{Query $\mathbf q\in\mathbb R^n$\\
Positive key $\mathbf k\in\mathbb R^n$\\
Temperature coefficient $\tau$\\
Image index $\mathbf I\in\mathbb N^n$
} 
\Output{The InfoNCE loss $l$}

    \BlankLine 
    $\mathbf q=\mathbf q/||\mathbf q||$\\
    $\mathbf k=\mathbf k/||\mathbf k||$\\
    $\mathbf W=\mathbf q\cdot\mathbf k^{\top}/\tau$~~\textcolor[rgb]{0.3, 0.7, 0.3}{\# $\mathbf W\in\mathbb R^{n\times n}$}\\
    Let sample mask $\theta_{ij} = \begin{cases} 
1 ~~~\text{if}~~~ \mathbf I_i=\mathbf I_j \\
0 ~~~\text{if}~~~ \mathbf I_i\ne\mathbf I_j
\end{cases}$\\
    $\mathbf e^{\texttt{all}}_i=\log\sum\limits_{j=1}^n\exp\left(\mathbf W_{ij}\right)$~~\textcolor[rgb]{0.3, 0.7, 0.3}{\# $\mathbf e^{\texttt{all}}\in\mathbb R^{n}$}\\
    $\mathbf e^{\texttt{pos}}_i=\log\sum\limits_{j=1}^n\theta_{ij}\exp\left(\mathbf W_{ij}\right)$~~\textcolor[rgb]{0.3, 0.7, 0.3}{\# $\mathbf e^{\texttt{pos}}\in\mathbb R^{n}$}\\
    $l=\sum\limits_{i=1}^n\left(\mathbf e^{\texttt{all}}_i-\mathbf e^{\texttt{pos}}_i\right)/n$\\
    \Return{$l$}
      \caption{$\mathcal L_{\texttt{Info}}$ for Multi Positive Samples}
      \label{algo_appendix:infonce} 
 \end{algorithm}
\makeatletter
\renewcommand{\thefigure}{C-\arabic{figure}}
\renewcommand{\thetable}{C-\arabic{table}}
\renewcommand{\theequation}{C-\arabic{equation}}
\renewcommand{\thealgocf}{C-\arabic{algocf}}
\makeatother
\setcounter{figure}{0}
\setcounter{table}{0}
\setcounter{equation}{0}
\setcounter{algocf}{0}

\section{Training Details}
\label{appendix:training_details}

\paragraph{Sphere Tokenizer.}
The training is conducted on a single NVIDIA A800 80GB GPU.
The hyperparameters are shown in Tab. \ref{appendix_tab:hyperparameters-sphere_tokenizer}.
We use a cosine learning rate scheduler during training.
To improve the robustness of the model, we introduce two data augmentation techniques: mixup with parameters \texttt{ratio=0.3} and \texttt{beta=0.3}, and the random rotation augmentation (maximum rotation angle: $5^\circ$) for position condition mentioned in our paper \textsection \ref{sec:method-sphere_tokenizer}.
To make the model focus more on visual brain regions, we only compute the loss for visual voxels.
\begin{table}[h]
    \centering
    \begin{tabular}{c|c}
    
        \specialrule{1.5pt}{2pt}{2pt}
        
        \textbf{Hyperparameters} & \textbf{Value} \\
        
        \specialrule{0.5pt}{0pt}{0pt}

        encoder channels & [64, 128, 256] \\
        dncoder channels & [32, 64, 128] \\
        hidden channels & 32 \\
        encoder ResNet blocks per down layer & 4 \\
        decoder ResNet blocks per down layer & 2 \\
        
        \specialrule{0.5pt}{0pt}{0pt}

        batch size & 32 \\
        learning rate & 4.0e-5 \\
        weight decay & 0.05 \\
        max gradient norm & 0.1 \\
        epoch & 80 \\
        
       \specialrule{1.5pt}{2pt}{2pt}
        
    \end{tabular}
    \caption{
    Hyperparameters for training sphere tokenizer.
    }
    \label{appendix_tab:hyperparameters-sphere_tokenizer}
    
\end{table}

\paragraph{fMRI Encoder.}
The training is conducted on a single NVIDIA A800 80GB GPU.
The hyperparameters are shown in Tab. \ref{appendix_tab:hyperparameters-sphere_tokenizer}.
We use a cosine learning rate scheduler during training.
To improve the model’s generalization ability, we applied augmentation to the images.
Each training image has a 50\% chance of being randomly horizontally flipped. 
The color properties are perturbed as follows: brightness, contrast, and saturation with a maximum perturbation of 0.2, and hue with a maximum perturbation of 0.1. 
Then the image is randomly rotated by up to $30^\circ$ and scaled within the range of [0.8, 1.0].
The temperature coefficient of $\mathcal L_{\texttt{biInfo}}$ is 0.1.
To avoid the influence of outliers, the CLIP embeddings are clamped to the range [-1.5, 1.5].
\begin{table}[h]
    \centering
    \begin{tabular}{c|c}
    
        \specialrule{1.5pt}{2pt}{2pt}
        
        \textbf{Hyperparameters} & \textbf{Value} \\

        \specialrule{0.5pt}{0pt}{0pt}

        embedding dim & 1024 \\
        MLP ratio & 4 \\
        depth & 24 \\
        num heads & 16 \\
        projection dropout & 0.5 \\
        
        \specialrule{0.5pt}{0pt}{0pt}

        batch size & 64 \\
        learning rate & 5.0e-4 \\
        weight decay & 0.05 \\
        max gradient norm & 0.5 \\
        epoch & 30 \\
        
       \specialrule{1.5pt}{2pt}{2pt}
        
    \end{tabular}
    \caption{
    Hyperparameters for training fMRI encoder.
    }
    \label{appendix_tab:hyperparamete-fmri_encoder}
    
\end{table}
\makeatletter
\renewcommand{\thefigure}{D-\arabic{figure}}
\renewcommand{\thetable}{D-\arabic{table}}
\renewcommand{\theequation}{D-\arabic{equation}}
\renewcommand{\thealgocf}{D-\arabic{algocf}}
\makeatother
\setcounter{figure}{0}
\setcounter{table}{0}
\setcounter{equation}{0}
\setcounter{algocf}{0}

\begin{table*}
    \centering
    \resizebox{\linewidth}{!}{
    \begin{tabular}{lcccccccc}
    
        \specialrule{1.5pt}{2pt}{2pt}

        \multirow{2.5}{*}{Inference} & 
        \multicolumn{4}{c}{Low-Level} &
        \multicolumn{4}{c}{High-Level}
        \\ 
        \cmidrule(lr){2-5} \cmidrule(lr){6-9} & 
        \texttt{PixCorr} $\uparrow$ &
        \texttt{SSIM} $\uparrow$ &
        \texttt{AlexNet}(\texttt{2}) $\uparrow$ &
        \texttt{AlexNet}(\texttt{5}) $\uparrow$ &
        \texttt{Inception} $\uparrow$ &
        \texttt{CLIP} $\uparrow$ &
        \texttt{EffNet-B} $\downarrow$ &
        \texttt{SwAV} $\downarrow$ \\
        
        \specialrule{0.5pt}{2pt}{2pt}

       subj01 &
       0.172 &
       0.314 &
       78.6\% &
       88.7\% &
       84.8\% &
       88.9\% &
       0.736 &
       0.396 \\

       subj02 &
       0.167 &
       0.302 &
       77.7\% &
       89.0\% &
       85.9\% &
       88.2\% &
       0.733 &
       0.394 \\

       subj05 &
       0.163 &
       0.305 &
       78.6\% &
       90.1\% &
       86.4\% &
       89.6\% &
       0.717 &
       0.393 \\

       subj07 &
       0.157 &
       0.298 &
       78.0\% &
       88.3\% &
       83.2\% &
       86.7\% &
       0.746 &
       0.409 \\

       \specialrule{0.5pt}{2pt}{2pt}
       
       Average &
       0.165 &
       0.305 &
       78.2\% &
       89.0\% &
       85.1\% &
       88.3\% &
       0.733 &
       0.398 \\
       
       \specialrule{1.5pt}{2pt}{2pt}
        
    \end{tabular}
    }
    \caption{
    Quantitative results for each subject on NSD \cite{allen2022massive} \texttt{test}.
    }
    \label{tab:details}
    
\end{table*}

\section{More Results}
\label{appendix:more_results}

\begin{figure*}[t]
    \centering
    \includegraphics[width=\textwidth]{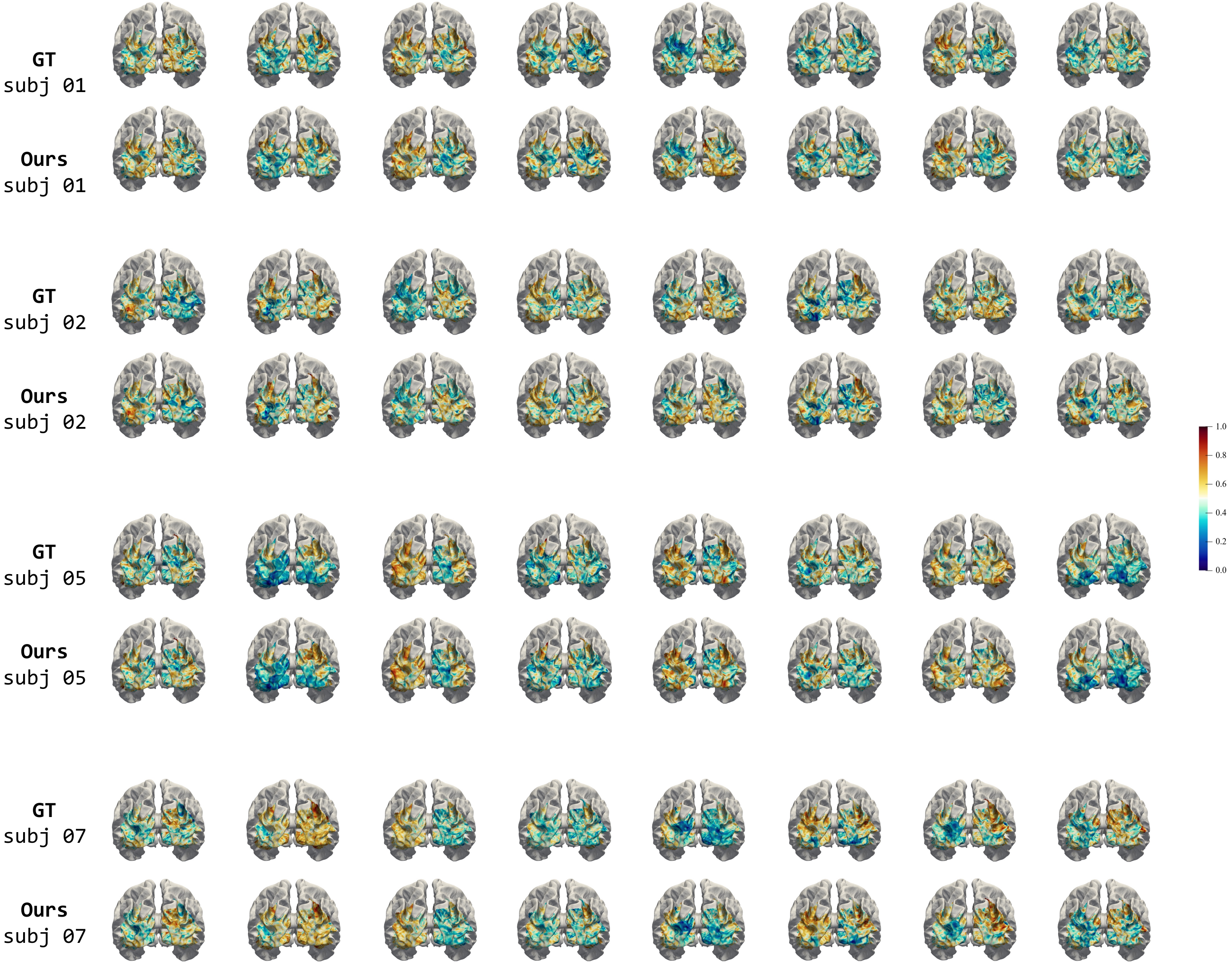}
    \caption{More fMRI vision voxels reconstruction results using the sphere tokenizer on the NSD \cite{allen2022massive} \texttt{test}.}
    \label{fig_appendix:fmri_rec}
\end{figure*}

\paragraph{Sphere Tokenizer.}
We present more reconstruction results of the sphere tokenizer autoencoder in Fig. \ref{fig_appendix:fmri_rec}, which demonstrates its effectiveness.

\paragraph{Comparison to Previous Work.}
We report the quantitative evaluation metrics in Tab. \ref{tab:compare}.
The results for Mind-Vis \cite{chen2023seeing} and MindEye \cite{scotti2024reconstructing} are cited from the report in \cite{quan2024psychometry}, while the result for UMBRAE \cite{xia2025umbrae} is from the report in \cite{shen2025neuro}.
The authors of NeuroPictor \cite{huo2025neuropictor} fine-tune the model for each subject to achieve higher performance. 
However, to ensure a fair comparison with other methods, we cite and report the version w/o fine-tuning.
The remaining methods are cited from their respective original papers.

\begin{figure*}[t]
    \centering
    \includegraphics[width=\textwidth]{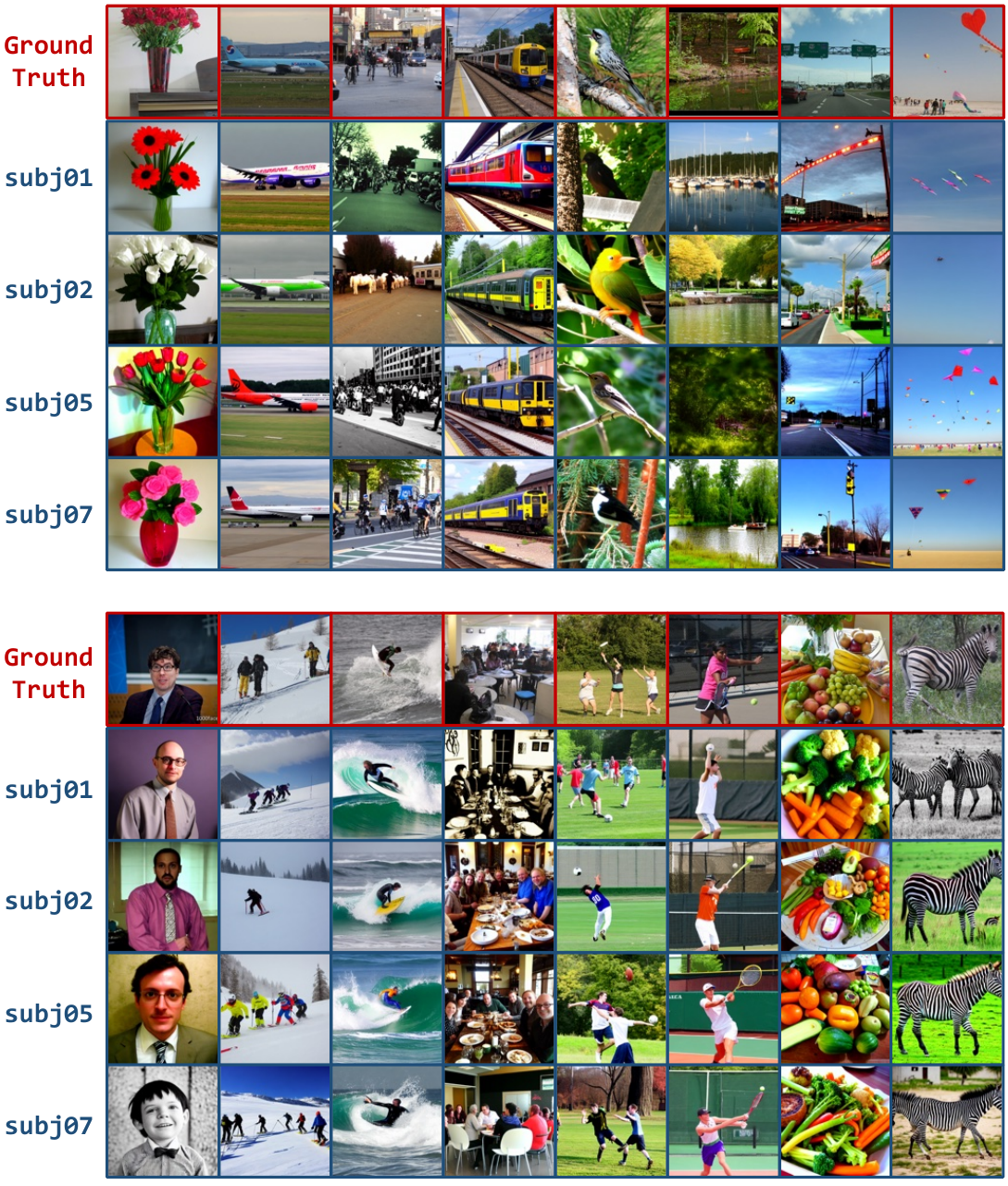}
    \caption{More fMRI-image reconstruction results on the NSD \cite{allen2022massive} \texttt{test}.}
    \label{fig_appendix:image_more_1}
\end{figure*}

\begin{figure*}[t]
    \centering
    \includegraphics[width=\textwidth]{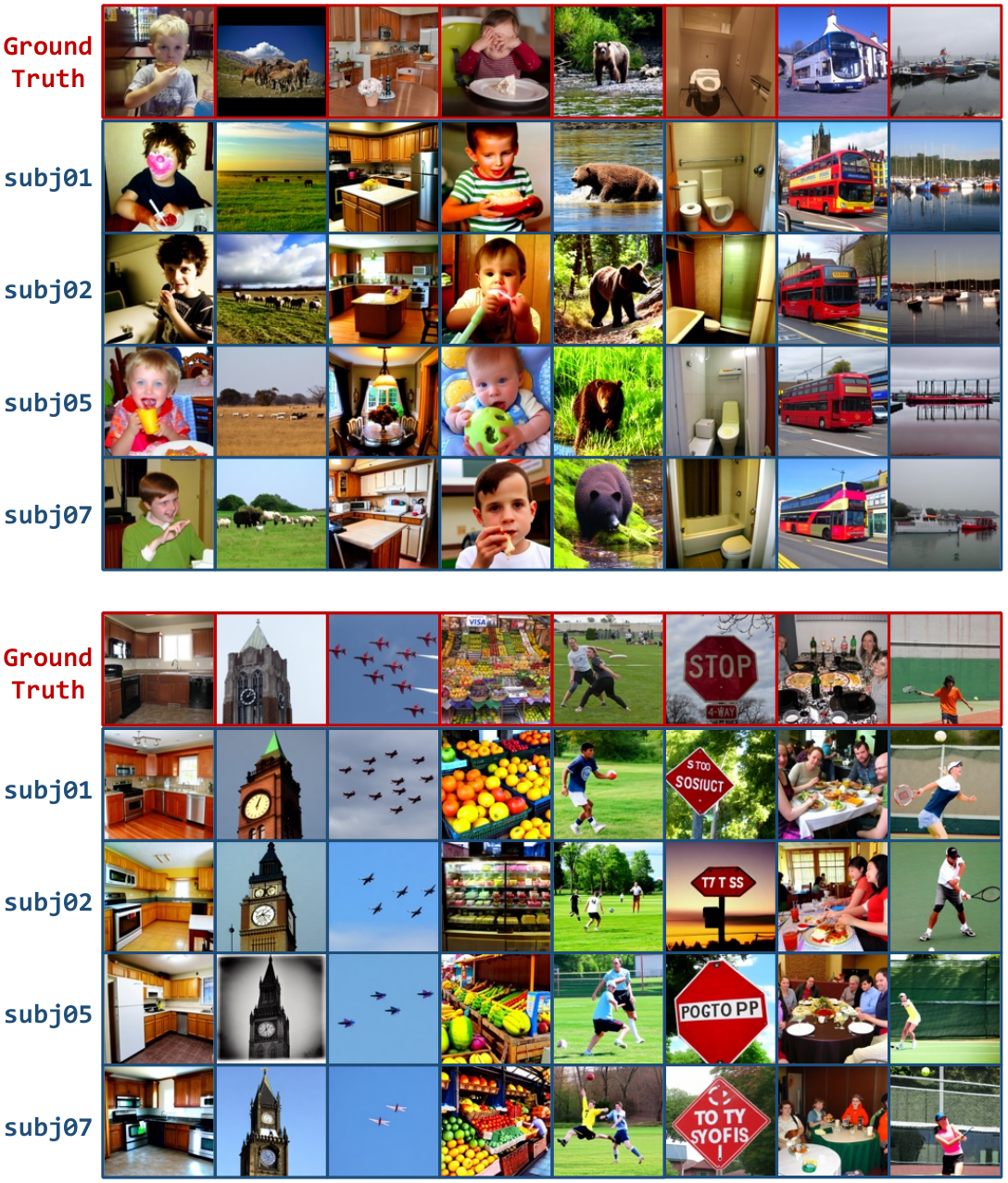}
    \caption{More fMRI-image reconstruction results on the NSD \cite{allen2022massive} \texttt{test}.}
    \label{fig_appendix:image_more_2}
\end{figure*}

\begin{figure*}[t]
    \centering
    \includegraphics[width=\textwidth]{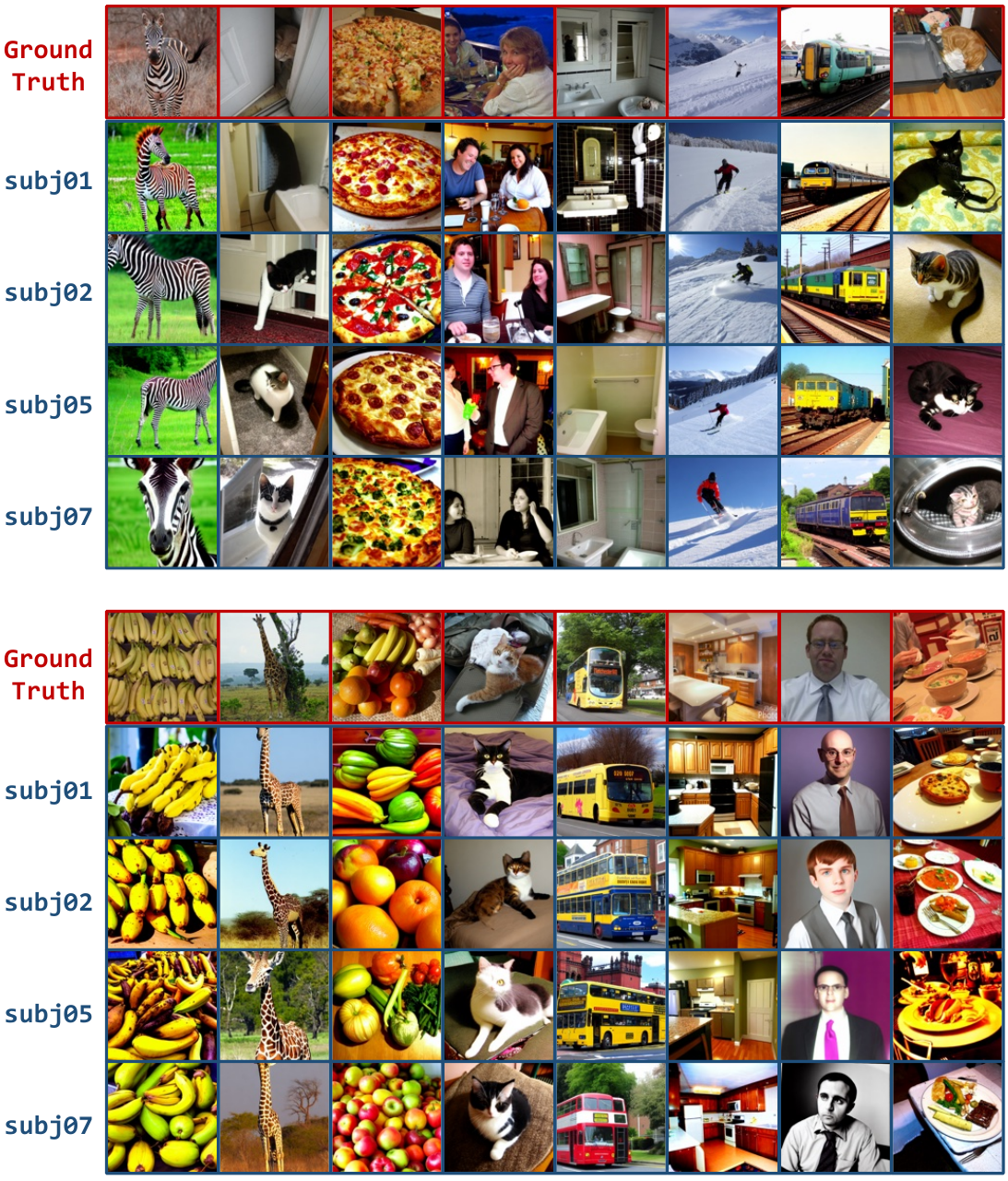}
    \caption{More fMRI-image reconstruction results on the NSD \cite{allen2022massive} \texttt{test}.}
    \label{fig_appendix:image_more_3}
\end{figure*}

\paragraph{Quantitative and Qualitative Results.}
We present the quantitative results on each subject of NSD \cite{allen2022massive} \texttt{test} in Tab. \ref{tab:details}.
We also provide additional decoding results in Fig. \ref{fig_appendix:image_more_1}, \ref{fig_appendix:image_more_2}, and \ref{fig_appendix:image_more_3}.
Readers can download all reconstruction images results from here:
\href{https://huggingface.co/datasets/yusijin/sphere_tokenizer_results_NSD}{\texttt{https://huggingface.co/datasets/yusijin/
sphere\_tokenizer\_results\_NSD}}.
\section{Limitations and Future Work}
Although we propose a novel approach for vision brain decoding, there are limitations that should be acknowledged.

\paragraph{Low fMRI Resolution.}
Our model is trained on low fMRI resolution, constrained by the spherical resolution supported by the standard SphericalUNet.
Experimental results demonstrate that fMRI resolution has a significant impact on performance.
In the future, we plan to (1) use higher-resolution spherical convolutions and (2) adjust the architecture of the sphere tokenizer to more efficiently handle low-resolution fMRI data.

\paragraph{Constrained Dataset Size.}
Similar to most previous work, we have only validated our results on the NSD dataset.
Although current techniques can handle cross-subject decoding, they are far from achieving cross-dataset generalization.
To achieve more robust vision brain decoding, larger-scale datasets are required.

\paragraph{Challenges in Practical Application.}
Our model is based on fMRI, which requires stringent conditions and high costs for data collection. 
This limits the widespread adoption and application of brain decoding technology.
Some technologies that are more suitable for real-time brain activity recording, such as EEG \cite{liu2024eeg2video} and fNIRS \cite{yu2024attention, chen2025sss}, fall far short of fMRI decoding performance due to their low signal-to-noise ratio.
This highlights the need for more efficient methods to enable models to capture representations of brain activity.

\paragraph{Challenges in Neuroscience.}
The scientific community still has no definitive understanding of the detailed mechanisms behind the functioning of the human brain.
The development of brain decoding can provide novel perspectives on this, highlighting the significance of biological interpretability in brain decoding models.
We will conduct a deeper exploration of this.

\paragraph{Privacy and Security Considerations.}
Personal brain activity data is highly sensitive private information, so protecting data security is crucial.
For example, membership inference attacks \cite{yu2025icas} and model inversion attacks \cite{yu2024calor, zhuang2025stealthy, qiu2024closer} can cause serious privacy leaks during the inference stage of the model.
We will incorporate considerations of data security in the future.

\end{document}